\documentclass{llncs}
\ifdefined\pdfminorversion\pdfminorversion=7\fi
\usepackage[T1]{fontenc}
\usepackage{lmodern}
\usepackage{graphicx}
\usepackage{amsmath,amssymb}
\usepackage{booktabs,array}
\usepackage{algorithm,algorithmic}
\usepackage[table]{xcolor}
\usepackage{arydshln}
\usepackage{placeins}
\usepackage[hidelinks]{hyperref}
\hypersetup{
  pdftitle={LVRE: A Local Visual Reasoning and Editing Agent for Optical Chemical Structure Recognition},
  pdfauthor={Bowen Lin and Lei Fan},
  pdfsubject={Optical chemical structure recognition via local visual reasoning and reversible graph editing},
  pdfkeywords={Optical chemical structure recognition, Vision-language models, Local visual verification, Graph editing}
}
\graphicspath{{figures/}}
\definecolor{methodgray}{gray}{0.93}
\definecolor{groupgray}{gray}{0.96}
\newcommand{\ours}{\textnormal{\textsc{LVRE}}}
\newcommand{\GraphAcc}{\mathrm{Acc}_{\mathrm{graph}}}
\newcommand{\lead}[1]{\par\smallskip\noindent\textbf{#1}\enspace}
\newcommand{\best}[1]{\textbf{#1}}
\begin{document}
\title{LVRE: A Local Visual Reasoning and Editing Agent for Optical Chemical Structure Recognition}
\author{Bowen Lin\inst{1} \and Lei Fan\inst{2}}
\institute{Department of Chemistry, University of Minnesota, USA
\and Shanghai Jiao Tong University, Shanghai, China}
\maketitle

\begin{abstract}
Optical chemical structure recognition (OCSR) demands strict all-or-nothing graph fidelity, where a single misidentified bond order or formal charge makes the predicted graph differ from the reference structure. While vision-language models (VLMs) can reinspect ambiguous visual regions, unconstrained full-graph regeneration often introduces regressive errors into previously correct substructures. \textbf{We present \ours{}, a training-free visual agent that refines molecular graphs via targeted visual inspection and reversible transactional editing.} Rather than regenerating the entire graph sequence, \ours{} examines a high-resolution crop alongside the global diagram and local graph neighborhood, proposing modifications to named graph records that a deterministic controller checks. A controller executes each proposed modification as an isolated transaction, and a visual review module arbitrates which edits to commit or roll back. Across four foundation VLMs, \ours{} improves graph accuracy by 8.0--8.5 percentage points over single-pass recognition and by 3.0--3.3 points over fixed local review. In a paired evaluation of 1,000 diagrams using Gemini 3.8 Flash, \ours{} achieves 100 successful repairs while introducing only 15 regressive errors, in contrast to global reflection which yields 65 repairs against 30 introduced errors. On this paired set, \ours{} introduces half as many errors on initially correct predictions as global reflection.
\end{abstract}

\section{Introduction}
A molecular diagram can be recognized almost entirely correctly and yet remain biochemically invalid. Omitting a single stroke converts a double bond into a single bond, while misinterpreting a charge label or atom symbol alters the valence state and functional identity of the compound. Optical chemical structure recognition (OCSR) must therefore recover subtle micro-details alongside global topological connectivity. Early rule-based pipelines such as OSRA combine classical image processing with heuristic chemical grammar~\cite{osra}. More recent deep learning approaches cast OCSR as image-to-sequence or image-to-graph translation~\cite{decimer,img2mol,molscribe}. However, diagrams extracted from scientific literature continue to pose severe challenges due to diverse typographic conventions, condensed substituent notations, and varying scan resolutions~\cite{molrec}.

Frontier vision-language models (VLMs) present an intuitive avenue for addressing recognition ambiguity by enabling iterative inspection and visual reasoning. A model can actively request alternative crops, compare candidate hypotheses, or revise intermediate predictions. Guided visual search improves reasoning about high-resolution images~\cite{vstar}, while sampling and iterative feedback have been studied as approaches to improving model outputs~\cite{consistency,selfrefine}. 
Nonetheless, unconstrained self-correction is rarely monotonically beneficial. Empirical studies on intrinsic self-correction demonstrate that autoregressive models frequently overwrite correct intermediate solutions with hallucinatory errors~\cite{selfcorrect}. In the context of OCSR, this manifests as an \emph{all-or-nothing generation dilemma}: prompting a VLM to regenerate a full molecular graph or linear string (e.g., SMILES) to fix a localized bond defect can corrupt unrelated parts of the graph.

We address this failure mode by strictly constraining the spatial and symbolic scope of every proposed modification. When localized uncertainty or topological invalidity is flagged, the agent inspects a magnified crop containing the target bond and its immediate topological neighborhood, proposing revisions exclusively to identified atom and bond attributes. Crucially, the non-target components of the graph remain immutable. 
Because a syntactically plausible edit may still misinterpret ambiguous visual strokes, syntactic constraint alone does not guarantee semantic accuracy. \ours{} therefore implements a closed-loop \emph{transactional controller}: proposed modifications are treated as pending transactions, and a visual verification module cross-examines the local patch against the original diagram to decide whether to commit the change or roll back to the prior graph state.
Our proposed agent, \ours{} (Local Visual Reasoning and Editing), operates on initial graph predictions generated by a frozen backbone VLM. The architecture unifies initial parsing, localized crop examination, and comparative review within the same shared backbone without task-specific fine-tuning. The deterministic controller coordinates crop extraction, maintains transaction histories, and enforces strict call-budget bounds. The primary contributions of this work are summarized as follows:
\begin{itemize}
    \item \textbf{A Transactional Graph-Editing Paradigm:} We formulate post-hoc OCSR refinement as a local graph-diff operation rather than global sequence regeneration, limiting changes to the declared graph records.
    \item \textbf{Closed-Loop Reversible Verification:} We design a visual review and rollback mechanism that evaluates proposed local changes against original diagram crops, reducing damage relative to global reflection on the paired Gemini set.
    \item \textbf{Fine-Grained Error Dynamics \& Efficiency Analysis:} Across four major VLM families, \ours{} delivers consistent +8.0--8.5 percentage point improvements in graph accuracy. Through paired diagnostics across 1,000 challenging diagrams, we explicitly isolate successful repairs from regressive edits and show that early stopping reduces relative cost by approximately 25\%, with a 0.2 percentage point difference in observed accuracy for the default backbone.
\end{itemize}

\section{Related Work}
\lead{Optical chemical structure recognition.} OSRA extracts chemical structures through image processing and chemical interpretation~\cite{osra}. DECIMER uses transformers to predict a molecular sequence~\cite{decimer}, whereas Img2Mol predicts a molecular latent representation that is decoded into SMILES~\cite{img2mol}. MolScribe predicts atoms, bonds, and their geometry~\cite{molscribe}. MolNexTR combines convolutional and transformer features with chemical post-processing~\cite{molnextr}. Oldenhof et al. study atom-level recognition with limited supervision and introduce ChemExpert, which selects among recognizer outputs according to chemical validity~\cite{oldenhof}. GTR-CoT trains a model to generate an atom--bond sequence through graph traversal~\cite{gtr}. \ours{} instead refines an existing VLM prediction. It uses the image to inspect proposed changes and does not call a specialized recognizer during refinement.

\lead{Visual inspection with tools.} ReAct interleaves model reasoning with actions that obtain information from an environment~\cite{react}. ViperGPT composes visual operations through executable programs~\cite{vipergpt}. V* searches for relevant image details~\cite{vstar}, and Visual Sketchpad creates and uses visual annotations during reasoning~\cite{sketchpad}. These methods make visual observation part of the inference process. In \ours{}, a crop is accompanied by the current graph neighborhood so that the model can refer to the records implicated by the image evidence. Our fixed-region baseline has the same inspection and editing tools, but follows a predefined region order.

\lead{Revision and verification.} Self-consistency selects answers from sampled reasoning paths~\cite{consistency}. Self-Refine alternates feedback and revision~\cite{selfrefine}, while Reflexion stores verbal feedback for later trials~\cite{reflexion}. These methods use different sources and forms of feedback. Huang et al. show that intrinsic self-correction can fail~\cite{selfcorrect}, and ProCo checks key conditions before attempting a correction~\cite{proco}. Work on test-time computation further shows that the benefit of additional inference depends on its allocation~\cite{testcompute}. Our experiments examine this trade-off for exact molecular graphs, where correcting an initial error and corrupting an initially correct prediction have opposite effects on accuracy.

\section{The LVRE Visual Agent}
\subsection{Task and Agent State}
Given an image $I$ of a single molecular structure, the task is to predict an attributed graph $G=(V,E,A)$, where $V$ contains atoms, $E$ contains bonds, and $A$ records their chemical attributes and supported groups. A fixed VLM $f_\theta$ produces $G_0=f_\theta(I)$. At step $t$, the agent stores the current graph $G_t$, an issue queue $Q_t$, a transaction history $H_t$, and the remaining call budget. Graph records have stable identifiers, atom labels and coordinates, bond endpoints and types, and supported charge, isotope, direction, or group information. Coordinates locate the relevant image region but are not used to score structural correctness. Page-level detection and reaction-role assignment are outside the task.

\begin{figure}[!htbp]
\centering
\includegraphics[width=\textwidth]{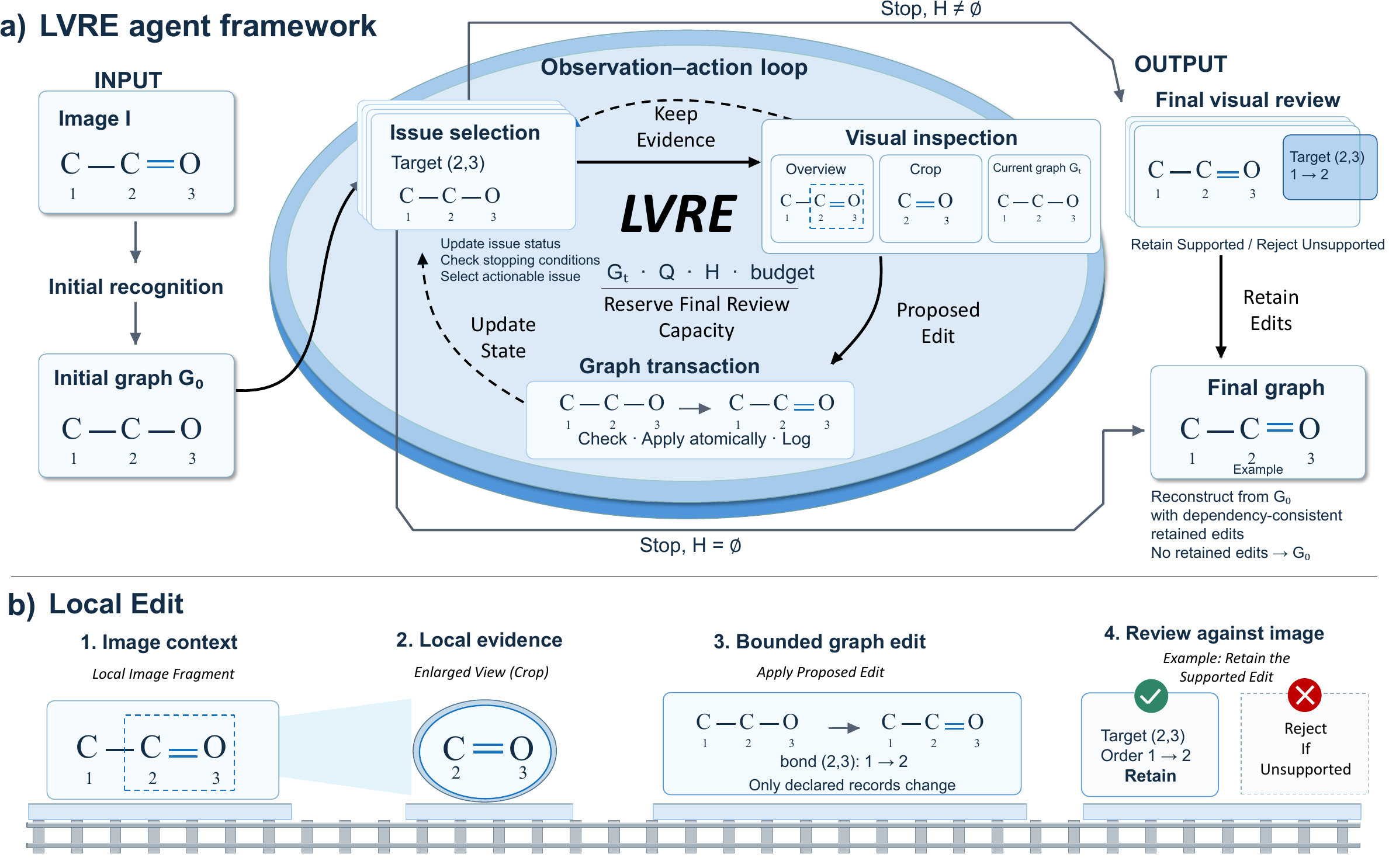}
\caption{\textbf{Overview of \ours{} and a schematic bond-order correction.} (a) The default five-call configuration reserves one call for final review. The agent inspects image crops and applies checked graph edits. The output retains the edits accepted at review, or returns $G_0$ if no edits remain. The double-bond output illustrates an accepted correction, not the empty-history path. (b) Image evidence supports changing bond $(2,3)$ from single to double. Stable atom identifiers link the views. Surrounding connections and hydrogens are omitted.}
\label{fig:framework}
\end{figure}

Each inspection concerns a named graph object and an ambiguity in its visual interpretation. The model proposes an edit to specified records, leaves the graph unchanged, or reports insufficient evidence. Text instructions describe the task and provide graph records and competing interpretations; structured text responses specify proposed edits and review decisions. The experiments do not isolate the contribution of the language modality. The controller records accepted transactions so that final review can revisit the changes without asking for another full graph prediction. Table~\ref{tab:notation} lists the state and edit notation, and Fig.~\ref{fig:framework} illustrates a bond-order correction. Reference checks establish that an edit can be executed, not that it agrees with the diagram.

\begin{table}[!htbp]
\caption{Agent state and edit representation in LVRE. Stable record identifiers link visual inspection, graph edits, and final review.}
\label{tab:notation}
\centering\small
\renewcommand{\arraystretch}{1.13}
\begin{tabular}{@{}p{2.3cm}p{1.55cm}p{\dimexpr\textwidth-3.85cm-4\tabcolsep\relax}@{}}
\toprule
\textbf{Role} & \textbf{Object} & \textbf{Operational meaning}\\
\midrule
Observation & $q_t, r_t$ & Localized issue and its image region. Includes target IDs and competing interpretations.\\
& $o_t$ & Overview, enlarged crop, and graph neighborhood with boundary connections.\\
\midrule
Modification & $\tau_t, S_t$ & Transaction and its declared set of affected record IDs.\\
& $h(G_t,\tau_t)$ & Transaction checks, including preconditions and edit-scope consistency.\\
\midrule
Acceptance & $H, z_j$ & Ordered change history and retain/reject decisions after visual review.\\
Budget & $K$ & Maximum calls, including initialization and any reserved final review.\\
\bottomrule
\end{tabular}
\end{table}

\subsection{Local Inspection and Graph Editing}
\lead{Issue selection and inspection.} After initialization, deterministic checks report duplicate IDs, missing endpoints, inconsistent fields, and invalid enumeration values. The VLM reads these diagnostics together with the image and graph, then returns up to two prioritized issues in the default configuration. An issue specifies a region, target IDs, and competing interpretations. Structural anomalies take priority over other localized ambiguities. For each issue, the model receives a full-image overview marking the region, an enlarged crop, and the graph neighborhood with its boundary connections. The surrounding atoms help identify where a short stroke or label belongs. A poorly localized crop allows one predefined margin expansion. Any further model judgment consumes another call. Crops use ordinary interpolation. The model's confidence estimate is not treated as a calibrated probability.

\lead{Transaction scope.} A proposed transaction $\tau_t$ records its base version, before-value preconditions, operations $\Delta_t$, and supporting image region. Operations can change labels, attributes, bond types or directions, add or delete records, reconnect endpoints, and update supported groups. Let $\mathcal R(G)$ be the mapping from stable IDs to atom, bond, and group records. The declared edit set $S_t$ includes all affected IDs, including new records and incident bonds handled when an atom is deleted. Define $U_t$ as the union of record IDs before and after the proposed edit. Records outside $S_t$ must remain unchanged:
\begin{equation}
 \mathcal R(\widetilde G_{t+1})\big|_{U_t\setminus S_t}
 =\mathcal R(G_t)\big|_{U_t\setminus S_t}.
 \label{eq:locality}
\end{equation}
This is a record-level scope constraint. It does not by itself bound the size of $S_t$. The proposed scope is tied to the selected issue and its supporting image region, and conflicts at its boundary cause the transaction to be rejected. Changing a bond order, for example, cannot alter an atom record outside the declared set. Within an affected record, the operations specify the fields to change.

\lead{Atomic execution.} The controller checks the base version, target references, and before-values against the current graph. It also checks that the proposed operations preserve valid references and satisfy Eq.~\eqref{eq:locality}. Let $h(G_t,\tau_t)$ indicate that these transaction checks pass. The update is
\begin{equation}
 G_{t+1}=\widetilde G_{t+1}=
 \begin{cases}
 \operatorname{Apply}(G_t,\Delta_t), & h(G_t,\tau_t)=1,\\
 G_t, & h(G_t,\tau_t)=0.
 \end{cases}
 \label{eq:apply}
\end{equation}
All operations succeed together or leave the graph unchanged. Deleting an atom requires explicit treatment of its incident bonds, and a failed reference check cannot leave a partial edit. The history stores before and after values for each applied transaction. These checks do not test visual correctness. An incorrect atom label may pass every reference check, and an unusual group should not be replaced merely because a chemistry tool finds it implausible.

\subsection{Visual Review and Budgeted Control}
\lead{Final review.} When edits exist and review is enabled, one final model call receives the original image, relevant crops, initial graph, and change list. It returns a retain/reject decision $z_j\in\{0,1\}$ for each transaction in $H$. Edits that cannot be separated because of their dependencies are reviewed together. The controller uses the stored before and after values to produce a graph containing the retained changes. A review decision does not override record preconditions or permit a dangling reference. If no changes remain, the output is $G_0$. The review response contains decisions about existing edits, not a newly generated molecular graph.

\lead{Call allocation.} The default cap of five model calls covers initialization, issue localization, up to two local inspections, and final review when edits exist. The controller reserves a review call before further inspection. It stops when no actionable issue remains, the evidence is insufficient, conflicting edits recur without new evidence, or the inspection budget is exhausted. The three-call configuration omits the separate final-review call. The seven-call configuration permits further inspection. Comparisons between these settings therefore change both the available stages and the call cap. Every model request, including a format retry, consumes budget. Algorithm~\ref{alg:controller} summarizes the procedure.

\begin{algorithm}[!htbp]
\caption{LVRE with a model-call cap}
\label{alg:controller}
\begin{algorithmic}[1]
\REQUIRE Image $I$, model $f_\theta$, cap $K\geq1$, review flag $v\in\{0,1\}$
\STATE Set remaining budget $b\leftarrow K$, $H\leftarrow\emptyset$
\STATE Generate and validate $G_0$. Set $G\leftarrow G_0$
\STATE Deduct every model request, including retries, from $b$
\IF{no usable graph or $b=0$}
\RETURN current valid output or protocol-compatible failure
\ENDIF
\STATE Localize issues $Q$ using $I$, $G_0$, and diagnostics. Deduct requests from $b$
\WHILE{an actionable issue remains and $b>v$}
\STATE Select an issue and obtain its overview, crop, and graph neighborhood
\STATE Request an edit, keep decision, or insufficient-evidence decision
\STATE Deduct requests from $b$, preserving the reserved review call if $v=1$
\IF{the response is a transaction $\tau$ and its checks pass}
\STATE Apply $\tau$ atomically. Store its before and after values in $H$
\ENDIF
\STATE Update issue status and check stopping conditions
\ENDWHILE
\IF{$H\ne\emptyset$ and $v=1$}
\STATE Review the changes with the reserved call. Set $b\leftarrow b-1$
\STATE Update $G$ from $G_0$ and the accepted, dependency-consistent changes in $H$
\ENDIF
\RETURN Export $G$ in the evaluation representation
\end{algorithmic}
\end{algorithm}

\section{Experiments}
We compare recognition strategies within each model, then examine ablations, paired changes in correctness, and inference cost. Graph accuracy is the primary outcome. SMILES accuracy is reported separately because its evaluation population differs.

\subsection{Experimental Setup}
\lead{Benchmark and metrics.} We follow the CARBON graph and SMILES evaluation conventions of MolRecBench-Wild~\cite{molrec,benchmarkcode}. The reference release is dated 19 August 2026. Its full dataset contains 5,024 graph records and 2,392 SMILES-eligible records, compared with 5,029 records in the earlier paper snapshot. These figures describe the benchmark release. Graph evaluation checks attributes, connectivity, and supported groups, excluding coordinates. The default SMILES convention retains atom chirality but ignores double-bond cis/trans slash directions. Recognition failures remain in evaluation. Reference graphs and difficulty labels are not available to the agent. We analyze the 1,000-image paired sets separately from the recognition summaries below.

\lead{Repair and damage.} Paired analysis uses 1,000 images per model, with a shared initial prediction for each within-model strategy comparison. Let $N_0$ and $N_1$ be the initial wrong and correct counts, and let $N_{01}$ and $N_{10}$ count wrong-to-correct and correct-to-wrong transitions. Repair rate $R$, damage rate $D$, and the change in graph accuracy are
\begin{equation}
 R=\frac{N_{01}}{N_0},\qquad D=\frac{N_{10}}{N_1},\qquad
 \Delta\GraphAcc=\frac{N_{01}-N_{10}}{N_0+N_1}.
 \label{eq:metrics}
\end{equation}
The rates are fractions. Tables report accuracy as a percentage and differences in percentage points (pp). A repair requires the final graph to be entirely correct: fixing one bond while leaving another error does not qualify. Damage means that an initially correct graph becomes incorrect. Equation~\eqref{eq:metrics} describes the net change on the paired set on which the counts are collected.

\begin{table}[!htbp]
\caption{\textbf{Recognition accuracy (\%) on MolRecBench-Wild.} B1--B3 and \ours{} use a five-call cap. Bold indicates the highest accuracy within each model and metric. Gain rows give the difference between \ours{} and each baseline in percentage points.}
\label{tab:main}
\centering\small
\setlength{\tabcolsep}{3pt}\renewcommand{\arraystretch}{1.14}
\begin{tabular}{@{}lrrrrrrrr@{}}
\toprule
& \multicolumn{2}{c}{\textbf{\shortstack{Gemini\\3.8 Flash}}} & \multicolumn{2}{c}{\textbf{\shortstack{Qwen3.8-\\Max}}} & \multicolumn{2}{c}{\textbf{\shortstack{Claude\\Opus 4.8}}} & \multicolumn{2}{c}{\textbf{\shortstack{GPT-5.6\\Sol}}}\\
\cmidrule(lr){2-3}\cmidrule(lr){4-5}\cmidrule(lr){6-7}\cmidrule(l){8-9}
\textbf{Strategy} & Graph & SMILES & Graph & SMILES & Graph & SMILES & Graph & SMILES\\
\midrule
B0 Single pass & 40.0 & 72.0 & 33.0 & 65.0 & 36.0 & 68.0 & 44.0 & 76.0\\
B1 Sampling & 44.8 & 75.0 & 37.2 & 68.1 & 40.2 & 71.1 & 48.4 & 78.8\\
B2 Reflection & 43.5 & 74.2 & 36.4 & 67.5 & 40.0 & 71.0 & 47.6 & 78.2\\
B3 Fixed local & 45.2 & 75.4 & 38.0 & 68.8 & 41.0 & 71.8 & 49.0 & 79.4\\
\rowcolor{methodgray}
\ours{} & \best{48.5} & \best{78.0} & \best{41.0} & \best{71.0} & \best{44.0} & \best{74.0} & \best{52.0} & \best{81.5}\\
\midrule
Gain vs. B0 & +8.5 & +6.0 & +8.0 & +6.0 & +8.0 & +6.0 & +8.0 & +5.5\\
Gain vs. B1 & +3.7 & +3.0 & +3.8 & +2.9 & +3.8 & +2.9 & +3.6 & +2.7\\
Gain vs. B2 & +5.0 & +3.8 & +4.6 & +3.5 & +4.0 & +3.0 & +4.4 & +3.3\\
Gain vs. B3 & +3.3 & +2.6 & +3.0 & +2.2 & +3.0 & +2.2 & +3.0 & +2.1\\
\bottomrule
\end{tabular}
\end{table}

\lead{Models.} Gemini 3.8 Flash~\cite{gemini38} is the default backbone. We also evaluate Qwen3.8-Max~\cite{qwen38}, Claude Opus 4.8~\cite{opus48}, and GPT-5.6 Sol~\cite{gpt56}. Each accepts image and text input through an API, and a trajectory uses the same model throughout. Inference uses API defaults for sampling, output limits, and reasoning controls where applicable. These defaults do not imply equal internal reasoning budgets across providers. The task prompts, tool access, and call caps are determined by the recognition strategy. Detailed comparisons of strategy-level repair and damage counts, as well as costs, use Gemini 3.8 Flash.

\lead{Baselines.} \emph{B0: single pass} returns the initial graph. \emph{B1: sampling and selection} generates four independent candidates and uses one image-conditioned call to select among them. \emph{B2: global reflection} allows up to four full-image, full-graph revisions after initialization. \emph{B3: fixed local review} uses the same crop, editing, review, and stopping machinery as \ours{}, but visits regions in a predefined order rather than choosing localized issues. B1--B3 and \ours{} share a five-call cap in the main comparison. Candidate selection and stopping do not have access to reference answers. We separately report results at three- and seven-call caps.

\lead{Ablations and costs.} We replace issue selection with fixed regions, remove enlarged crops, replace bounded editing with graph rewriting, remove final review, or disable early stopping. B3 also serves as the region-selection ablation. Global rewriting retains local inspection and therefore differs from B2. Removing review also removes its possible call, so this ablation is interpreted jointly with paired outcomes and cost. For image $i$, let $n_i$ be the number of model requests and $c_{ij}$ the cost assigned to request $j$. The mean and relative costs are
\begin{equation}
 \overline C=\frac{1}{N}\sum_{i=1}^N\sum_{j=1}^{n_i} c_{ij},\qquad
 C_{\mathrm{rel}}=\frac{\overline C}{\overline C_{\mathrm{single}}}.
 \label{eq:cost}
\end{equation}
Cost accounting includes applicable image, text, output, reasoning, selection, and retry charges. Initialization is counted for each strategy even when paired comparisons reuse a cached initial prediction. The denominator is the same model's mean single-pass cost. All cost comparisons below use Gemini 3.8 Flash. A shared call cap does not imply equal expenditure, and normalized costs cannot establish absolute price rankings across providers.

\subsection{Recognition Performance}

Table~\ref{tab:main} reports higher graph and SMILES accuracy for \ours{} than for all four baselines on each model. The graph improvement over single pass is 8.0--8.5 pp. Sampling and global reflection also improve on single pass, but their gains are smaller. B3 is the strongest baseline, with graph accuracy 3.0--3.3 pp below \ours{}. Since B3 already uses local crops, graph edits, and final review, this comparison supports choosing regions according to identified issues rather than following a fixed order. SMILES accuracy improves by 5.5--6.0 pp over single pass and by 2.1--2.6 pp over B3.

\lead{Removing crop enlargement causes the largest drop.} In Table~\ref{tab:ablation}, removing the enlarged crop reduces graph accuracy by 3.4--3.9 pp across models. Replacing issue selection with fixed regions reduces it by 3.0--3.3 pp. Allowing full-graph rewriting after local inspection reduces accuracy by 2.8--3.0 pp, suggesting that the form of the update matters even when the model receives local evidence. Removing final review has a smaller effect of 1.2--1.7 pp. The paired analysis in Sect.~\ref{sec:behavior} examines how review changes repair and damage counts. Disabling early stopping slightly increases accuracy, at higher cost.

\begin{table}[!htbp]
\caption{\textbf{Ablation results.} Graph accuracy (\%) for four backbones and relative inference cost for Gemini 3.8 Flash, normalized to its single-pass cost. The full LVRE configuration is shaded. Bold indicates the highest accuracy in each model column and the lowest cost among the listed configurations.}
\label{tab:ablation}
\centering\small
\setlength{\tabcolsep}{5pt}\renewcommand{\arraystretch}{1.12}
\begin{tabular}{@{}lrrrrr@{}}
\toprule
& \multicolumn{4}{c}{\textbf{Graph accuracy} $\uparrow$} & \textbf{Cost} $\downarrow$\\
\cmidrule(lr){2-5}\cmidrule(l){6-6}
\textbf{Configuration} & \shortstack{Gemini\\3.8 Flash} & \shortstack{Qwen3.8-\\Max} & \shortstack{Claude\\Opus 4.8} & \shortstack{GPT-5.6\\Sol} & \shortstack{Gemini\\3.8 Flash}\\
\midrule
\rowcolor{methodgray}\ours{} & 48.5 & 41.0 & 44.0 & 52.0 & 3.6$\times$\\
\midrule
\multicolumn{6}{@{}l}{\emph{Observation and modification}}\\
\hdashline
Fixed regions & 45.2 & 38.0 & 41.0 & 49.0 & 3.6$\times$\\
No enlarged crop & 44.6 & 37.5 & 40.4 & 48.6 & 3.5$\times$\\
Global graph rewrite & 45.5 & 38.2 & 41.2 & 49.1 & 3.8$\times$\\
\addlinespace[2pt]
\multicolumn{6}{@{}l}{\emph{Acceptance and allocation}}\\
\hdashline
No final review & 46.8 & 39.8 & 42.5 & 50.5 & \best{2.8}$\times$\\
No early stopping & \best{48.7} & \best{41.3} & \best{44.1} & \best{52.2} & 4.8$\times$\\
\bottomrule
\end{tabular}
\end{table}

\lead{Comparison with specialized recognizers.} The benchmark repository reports SMILES accuracies of 62.29\% for MolScribe and 62.50\% for MolNexTR. MinerU.Chem GTR-VL-1.4.13 reaches 79.66\% graph accuracy and 93.02\% SMILES accuracy~\cite{benchmarkcode}. These are external reference results, not systems remeasured under our inference protocol. They show that improving the evaluated VLM strategies does not close the gap to a strong specialized recognizer. We do not compare their costs with those of \ours{}.

\subsection{Repair and Preservation}\label{sec:behavior}

\begin{figure}[!htbp]
\centering
\includegraphics[width=\textwidth]{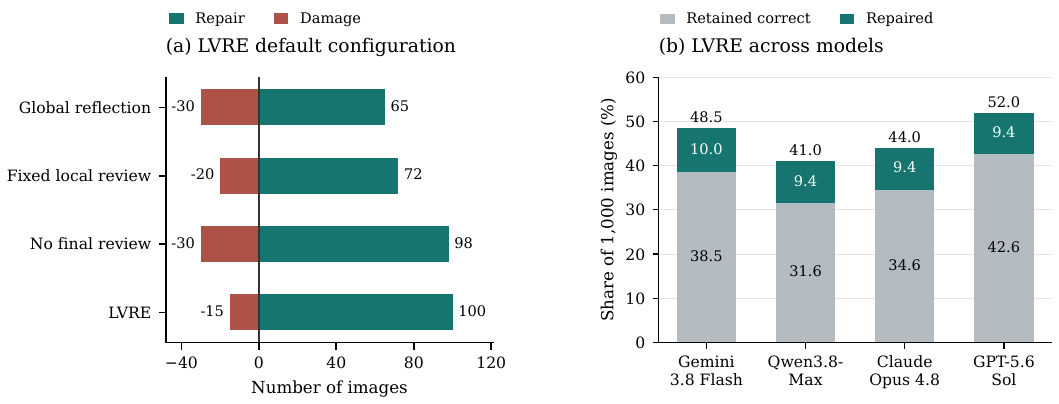}
\caption{\textbf{Changes in correctness on 1,000 images per model.} (a) All configurations use Gemini 3.8 Flash and share 400 initially correct and 600 initially incorrect predictions. Rightward bars count wrong-to-correct transitions, and leftward bars count correct-to-wrong transitions. (b) For \ours{}, stacked bars show predictions that remain correct and those corrected from an initially incorrect graph, as percentages of all 1,000 images.}
\label{fig:behavior}
\end{figure}

Figure~\ref{fig:behavior}(a) reports changes on the paired 1,000-image set for Gemini 3.8 Flash. \ours{} corrects 100 initially incorrect graphs and makes 15 initially correct graphs incorrect. Global reflection has 65 repairs and 30 cases of damage. The net difference is 50 correct predictions on this set: 35 from the higher repair count and 15 from the lower damage count. Against fixed local review, the corresponding differences are 28 and five, giving a net advantage of 33 predictions, or 3.3 pp on the paired set. These are aggregate count differences, not counts of individually traced review decisions.

\lead{Final review reduces damage more than it increases repairs.} On the same paired set, adding final review increases repairs from 98 to 100 and reduces the damage count from 30 to 15. The net gain is 17 correct predictions, or 1.7 pp on this set. Of this net gain, two predictions come from the increase in repairs and 15 from the reduction in damage. This pattern is consistent with review helping to preserve initially correct predictions. It does not isolate the effect of individual rollback decisions, since the ablation also changes the possible call allocation.

Across the four models, \ours{} repairs 14.0--16.8\% of initially incorrect graphs and damages 3.2--4.2\% of initially correct graphs. These conditional rates use different denominators from the percentages in Fig.~\ref{fig:behavior}(b), which refer to all 1,000 images. Every model has a positive net change, although none eliminates damage.

\subsection{Inference Budget and Cost}

\begin{figure}[!htbp]
\centering
\includegraphics[width=\textwidth]{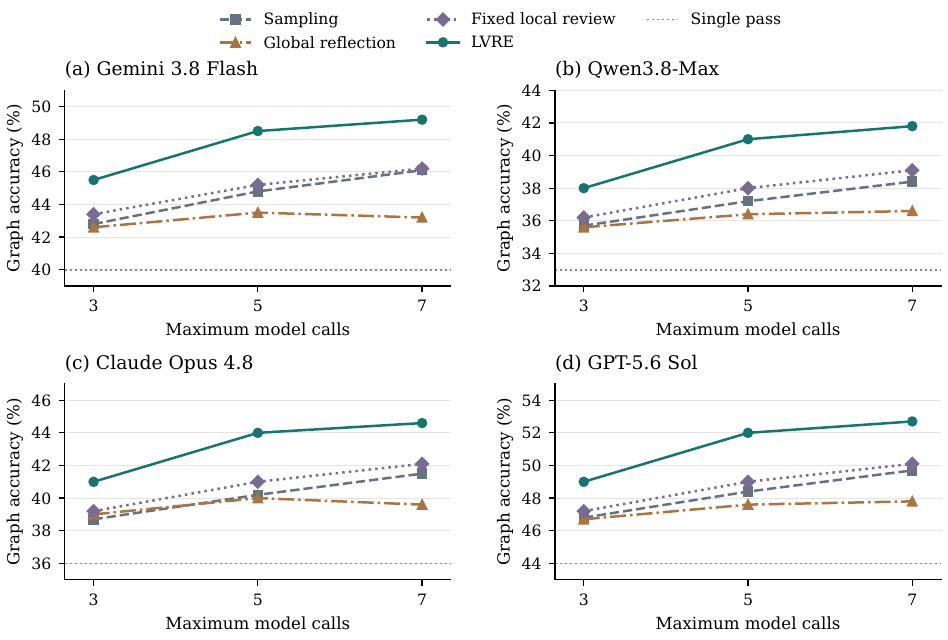}
\caption{\textbf{Graph accuracy at different model-call caps.} Each panel shows one backbone. Markers denote evaluated settings, and connecting lines are visual guides. Gray horizontal dotted lines show single-pass accuracy. \ours{} omits the separate final-review call at $K=3$ and permits it at $K=5$ and $K=7$.}
\label{fig:budget}
\end{figure}

\lead{Gains diminish beyond the five-call configuration.} In Fig.~\ref{fig:budget}, \ours{} has the highest graph accuracy at each evaluated call cap for every model. The three-call configuration gains 5.0--5.5 pp over single pass. The five-call configuration adds 3.0 pp, and the seven-call configuration adds a further 0.6--0.8 pp. The first increase also enables final review, so it is not solely an effect of more inspection calls. Global reflection is less consistent: moving from five to seven calls reduces accuracy by 0.3 pp for Gemini 3.8 Flash and 0.4 pp for Claude Opus 4.8, while slightly improving the other two models. Under the five-call cap, \ours{} uses an average of 3.5--3.7 calls across models.

\begin{figure}[!htbp]
\centering
\includegraphics[width=\textwidth]{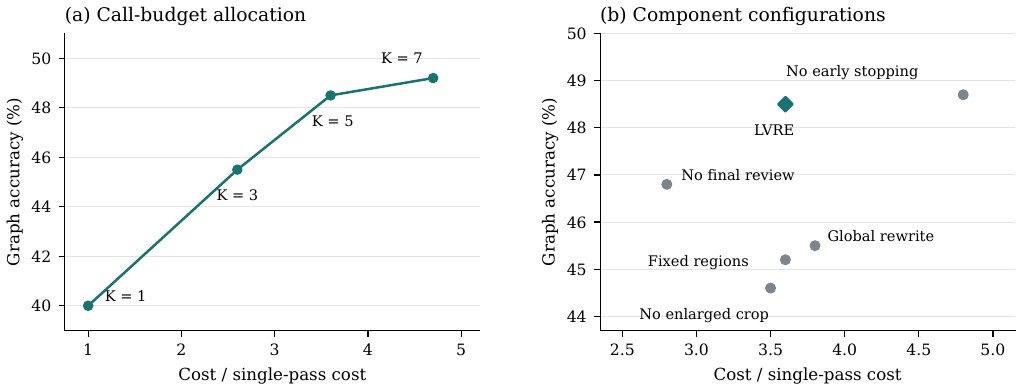}
\caption{\textbf{Graph accuracy versus relative inference cost with Gemini 3.8 Flash.} (a) Single-pass recognition ($K=1$) and \ours{} at call caps $K=3,5,7$. (b) \ours{} and its ablations. Cost is the mean cost per image divided by that of single-pass recognition. Each point represents an evaluated configuration.}
\label{fig:cost}
\end{figure}

\lead{Early stopping lowers cost.} For Gemini 3.8 Flash, moving from five to seven calls increases relative cost from 3.6 to 4.7 and graph accuracy by 0.7 pp (Fig.~\ref{fig:cost}). Disabling early stopping at the default cap raises cost from 3.6 to 4.8 for an observed gain of 0.2 pp. Using the reported costs, early stopping therefore saves approximately 25\%. The small accuracy difference is a point estimate, not evidence of a reproducible advantage for either stopping policy.

Other configurations have different cost profiles. Fixed regions and \ours{} both cost 3.6 times single pass at the reported precision, but differ by 3.3 pp in graph accuracy. Global graph rewriting costs 3.8 times single pass and reaches 45.5\% accuracy. Removing final review lowers relative cost to 2.8 and accuracy to 46.8\%. These comparisons apply to the default backbone.

\FloatBarrier
\section{Conclusion}
In this work, we address a critical vulnerability in VLM-based optical chemical structure recognition: the tendency of unconstrained self-correction to corrupt globally correct structures while attempting local fixes. We introduce \ours{}, a training-free agentic framework that replaces degradation-prone, whole-graph regeneration with localized visual inspection and transactional graph editing. By decoupling local high-resolution re-inspection from global graph state maintenance, \ours{} consistently improves molecular graph accuracy across four foundation VLMs, outperforming both standard single-pass inference and fixed local review baselines.
On the paired Gemini set, \ours{} repairs 100 initially incorrect graphs and makes 15 initially correct graphs incorrect, compared with 65 and 30, respectively, for global reflection. The reduction in damage in the review ablation is consistent with a benefit from final review, although the ablation also changes call allocation. Budget-aware early stopping reduces relative cost by approximately 25\%, with a small observed difference in accuracy in this experiment.

\FloatBarrier


\begin{thebibliography}{99}
\raggedright
\bibitem{osra}
Filippov, I.V., Nicklaus, M.C.: Optical structure recognition software to recover chemical information: OSRA---an open source solution. Journal of Chemical Information and Modeling \textbf{49}(3), 740--743 (2009). \doi{10.1021/ci800067r}
\bibitem{decimer}
Rajan, K., Zielesny, A., Steinbeck, C.: DECIMER 1.0: Deep learning for chemical image recognition using transformers. Journal of Cheminformatics \textbf{13}, 61 (2021). \doi{10.1186/s13321-021-00538-8}
\bibitem{img2mol}
Clevert, D.-A., Le, T., Winter, R., Montanari, F.: Img2Mol---accurate SMILES recognition from molecular graphical depictions. Chemical Science \textbf{12}, 14174--14181 (2021). \doi{10.1039/D1SC01839F}
\bibitem{molscribe}
Qian, Y., Guo, J., Tu, Z., Li, Z., Coley, C.W., Barzilay, R.: MolScribe: Robust molecular structure recognition with image-to-graph generation. Journal of Chemical Information and Modeling \textbf{63}(7), 1925--1934 (2023). \doi{10.1021/acs.jcim.2c01480}
\bibitem{molrec}
Yang, H., et al.: MolRecBench-Wild: A real-world benchmark for optical chemical structure recognition. arXiv:2605.05832 (2026). \doi{10.48550/arXiv.2605.05832}
\bibitem{consistency}
Wang, X., et al.: Self-consistency improves chain of thought reasoning in language models. In: International Conference on Learning Representations (2023). \url{https://arxiv.org/abs/2203.11171}
\bibitem{selfrefine}
Madaan, A., et al.: Self-Refine: Iterative refinement with self-feedback. In: Advances in Neural Information Processing Systems, vol. 36 (2023)
\bibitem{vstar}
Wu, P., Xie, S.: V*: Guided visual search as a core mechanism in multimodal LLMs. In: Proceedings of the IEEE/CVF Conference on Computer Vision and Pattern Recognition, pp. 13084--13094 (2024)
\bibitem{selfcorrect}
Huang, J., et al.: Large language models cannot self-correct reasoning yet. In: International Conference on Learning Representations (2024). \url{https://arxiv.org/abs/2310.01798}
\bibitem{molnextr}
Chen, Y., Leung, C.T., Huang, Y., Sun, J., Chen, H., Gao, H.: MolNexTR: A generalized deep learning model for molecular image recognition. Journal of Cheminformatics \textbf{16}, 141 (2024). \doi{10.1186/s13321-024-00926-w}
\bibitem{oldenhof}
Oldenhof, M., De Brouwer, E., Arany, A., Moreau, Y.: Atom-level optical chemical structure recognition with limited supervision. In: Proceedings of the IEEE/CVF Conference on Computer Vision and Pattern Recognition, pp. 17669--17678 (2024). \url{https://arxiv.org/abs/2404.01743}
\bibitem{gtr}
Wang, J., et al.: GTR-CoT: Graph traversal as visual chain of thought for molecular structure recognition. arXiv:2506.07553 (2025). \doi{10.48550/arXiv.2506.07553}
\bibitem{react}
Yao, S., Zhao, J., Yu, D., Du, N., Shafran, I., Narasimhan, K., Cao, Y.: ReAct: Synergizing reasoning and acting in language models. In: International Conference on Learning Representations (2023). \url{https://arxiv.org/abs/2210.03629}
\bibitem{vipergpt}
Sur\'is, D., Menon, S., Vondrick, C.: ViperGPT: Visual inference via Python execution for reasoning. In: Proceedings of the IEEE/CVF International Conference on Computer Vision (2023). \url{https://arxiv.org/abs/2303.08128}
\bibitem{sketchpad}
Hu, Y., Shi, W., Fu, X., Roth, D., Ostendorf, M., Zettlemoyer, L., Smith, N.A., Krishna, R.: Visual Sketchpad: Sketching as a visual chain of thought for multimodal language models. In: Advances in Neural Information Processing Systems, vol. 37 (2024)
\bibitem{reflexion}
Shinn, N., et al.: Reflexion: Language agents with verbal reinforcement learning. In: Advances in Neural Information Processing Systems, vol. 36 (2023). \url{https://arxiv.org/abs/2303.11366}
\bibitem{proco}
Wu, Z., Zeng, Q., Zhang, Z., Tan, Z., Shen, C., Jiang, M.: Large language models can self-correct with key condition verification. In: Proceedings of EMNLP, pp. 12846--12867 (2024). \doi{10.18653/v1/2024.emnlp-main.714}
\bibitem{testcompute}
Snell, C., Lee, J., Xu, K., Kumar, A.: Scaling LLM test-time compute optimally can be more effective than scaling parameters for reasoning. In: International Conference on Learning Representations (2025). \url{https://openreview.net/forum?id=4FWAwZtd2n}
\bibitem{benchmarkcode}
OpenDataLab: MolRecBench-Wild: Official benchmark and evaluation repository. Release 2026-08-19. \url{https://github.com/opendatalab/MolRecBench-Wild}. Accessed 14 September 2026
\bibitem{gemini38}
Google: Gemini 3.8 Flash. Gemini API model documentation (2026). \url{https://ai.google.dev/gemini-api/docs/models/gemini-3.8-flash}. Accessed 15 September 2026
\bibitem{qwen38}
Alibaba Cloud: Qwen3.8-Max. Model Studio model documentation (2026). \url{https://www.alibabacloud.com/help/en/model-studio/qwen3-8-max}. Accessed 15 September 2026
\bibitem{opus48}
Anthropic: \href{https://www-cdn.anthropic.com/0b4915911bb0d19eca5b5ee635c80fef830a37ea.pdf}{Claude Opus 4.8 System Card} (2026). Accessed 15 September 2026
\bibitem{gpt56}
OpenAI: GPT-5.6 Sol. API model documentation (2026). \url{https://developers.openai.com/api/docs/models/gpt-5.6-sol}. Accessed 15 September 2026
\end{thebibliography}
\end{document}